\documentclass[letterpaper, 10 pt, conference]{ieeeconf}%

\IEEEoverridecommandlockouts                              %

\usepackage{amsmath}
\usepackage{mathtools}
\usepackage{booktabs}
\usepackage{float}
\usepackage{xspace}
\usepackage{cite}
\usepackage{graphicx}
\graphicspath{{./figures/}}
\usepackage{caption}
\usepackage{subcaption}
\usepackage[inline]{asymptote}
\usepackage{tikz}
\usepackage{comment}
\usepackage{xcolor}
\usepackage{hyperref}
\hypersetup{
    colorlinks=true,
    urlcolor=magenta
    }
    
\PassOptionsToPackage{hyphens}{url}\usepackage{hyperref}

\usetikzlibrary{bayesnet}
\usetikzlibrary{arrows, positioning}

\newcommand{\SEthree}{\ensuremath{\mathrm{SE}(3)}\xspace}
\newcommand{\residual}{\mathbf{r}}
\DeclareMathOperator*{\argmin}{arg\,min}
\newcommand{\ra}[1]{\renewcommand{\arraystretch}{#1}}
\newcommand{\States}{\mathcal{X}}
\newcommand{\state}{\mathrm{\mathbf{x}}}

\newcommand{\vel}{\mathbf{v}}
\newcommand{\bias}{\mathbf{b}}

\newcommand{\World}{\mathtt{W}}
\newcommand{\Imu}{\mathtt{B}}
\newcommand{\Camera}{\mathtt{C}}

\newcommand{\imu}{\mathtt{\tiny{B}}}
\newcommand{\camera}{\mathtt{\tiny{C}}}

\newcommand{\R}{\mathtt{R}}
\newcommand{\trans}{\mathbf{t}}
\newcommand{\T}{\mathtt{T}}
\newcommand{\transpose}{\mathsf{T}}
\newcommand{\Identity}{\mathtt{I}}
\newcommand{\Homography}{\mathtt{H}}

\newcommand{\px}{\mathbf{u}} %
\newcommand{\pt}{\mathbf{p}} %
\newcommand{\depth}{d}

\newcommand{\n}{\mathbf{n}} %

\newcommand{\mRot}{\Delta\tilde\R_{ij}}

\usepackage{array}
\newcommand{\PreserveBackslash}[1]{\let\temp=\\#1\let\\=\temp}
\newcolumntype{C}[1]{>{\PreserveBackslash\centering}p{#1}}
\newcolumntype{L}[1]{>{\PreserveBackslash\raggedright}p{#1}}

\title{\LARGE \bf
RP-VIO: Robust Plane-based Visual-Inertial Odometry for Dynamic Environments
}

\begin{document}

\global\csname @topnum\endcsname 0
\global\csname @botnum\endcsname 0

\bstctlcite{Force_Etal}

\author{Karnik Ram, Chaitanya Kharyal, Sudarshan S. Harithas, K. Madhava Krishna
\thanks{All authors are with the Robotics Research Center at IIIT Hyderabad, India. Correspondence email: \texttt{karnikram@gmail.com}. The authors thank the anonymous reviewers for helpful comments, and MathWorks India Hyderabad for generous financial support.}\\
\thanks{$^\dagger$Project page: \url{https://rebrand.ly/rp-vio}}}

\newcommand\copyrighttext{%
  \footnotesize \textcopyright 2021 IEEE. Personal use of this material is permitted.
  Permission from IEEE must be obtained for all other uses, in any current or future
  media, including reprinting/republishing this material for advertising or promotional
  purposes, creating new collective works, for resale or redistribution to servers or
  lists, or reuse of any copyrighted component of this work in other works.
  eCF ID: ras.IROS21.223.71579967}
\newcommand\copyrightnotice{%
\begin{tikzpicture}[remember picture,overlay]
\node[anchor=south,yshift=10pt] at (current page.south) {\fbox{\parbox{\dimexpr\textwidth-\fboxsep-\fboxrule\relax}{\copyrighttext}}};
\end{tikzpicture}%
}

\maketitle
\hypersetup{urlcolor=black}
\copyrightnotice
\thispagestyle{empty}
\pagestyle{empty}

\begin{abstract}
    Modern visual-inertial navigation systems (VINS) are faced with a critical challenge in real-world deployment: they need to operate reliably and robustly in highly dynamic environments. Current best solutions merely filter dynamic objects as outliers based on the semantics of the object category. Such an approach does not scale as it requires semantic classifiers to encompass all possibly-moving object classes; this is hard to define, let alone deploy. On the other hand, many real-world environments exhibit strong structural regularities in the form of planes such as walls and ground surfaces, which are also crucially static. We present RP-VIO, a monocular visual-inertial odometry system that leverages the simple geometry of these planes for improved robustness and accuracy in challenging dynamic environments. Since existing datasets have a limited number of dynamic elements, we also present a highly-dynamic, photorealistic synthetic dataset for a more effective evaluation of the capabilities of modern VINS systems. We evaluate our approach on this dataset, and three diverse sequences from standard datasets including two real-world dynamic sequences and show a significant improvement in robustness and accuracy over a state-of-the-art monocular visual-inertial odometry system. We also show in simulation an improvement over a simple dynamic-features masking approach. Our code and dataset are publicly available$^\dagger$.
\end{abstract}

\section{Introduction}
    The visual-inertial navigation systems (VINS) of today are cheap, compact, and provide geometry and pose estimates in real-time with centimeter-level accuracy. VINS are increasingly being used in mobile robot navigation, virtual reality, and augmented reality applications~\cite{eth-nav, oculus, hololens}. Cameras and inertial measurement units (IMUs) in VINS complement each other: IMUs resolve the scale factor ambiguity with monocular cameras, while cameras render the unobservable IMU biases and intrinsics observable. Yet, the approach has some limitations. Apart from the additional hardware that needs to be accurately synchronized and calibrated, the system needs to perform sufficient rotation and acceleration motions to keep the gravity and scale observable \cite{vins-wheels}. For extended operation, VINS also require online calibration where degenerate trajectories may render the extrinsics and intrinsics unobservable~\cite{degenerate-ex, degenerate-in}.

\begin{figure}[t!]
    \centering
    \includegraphics[scale=0.5]{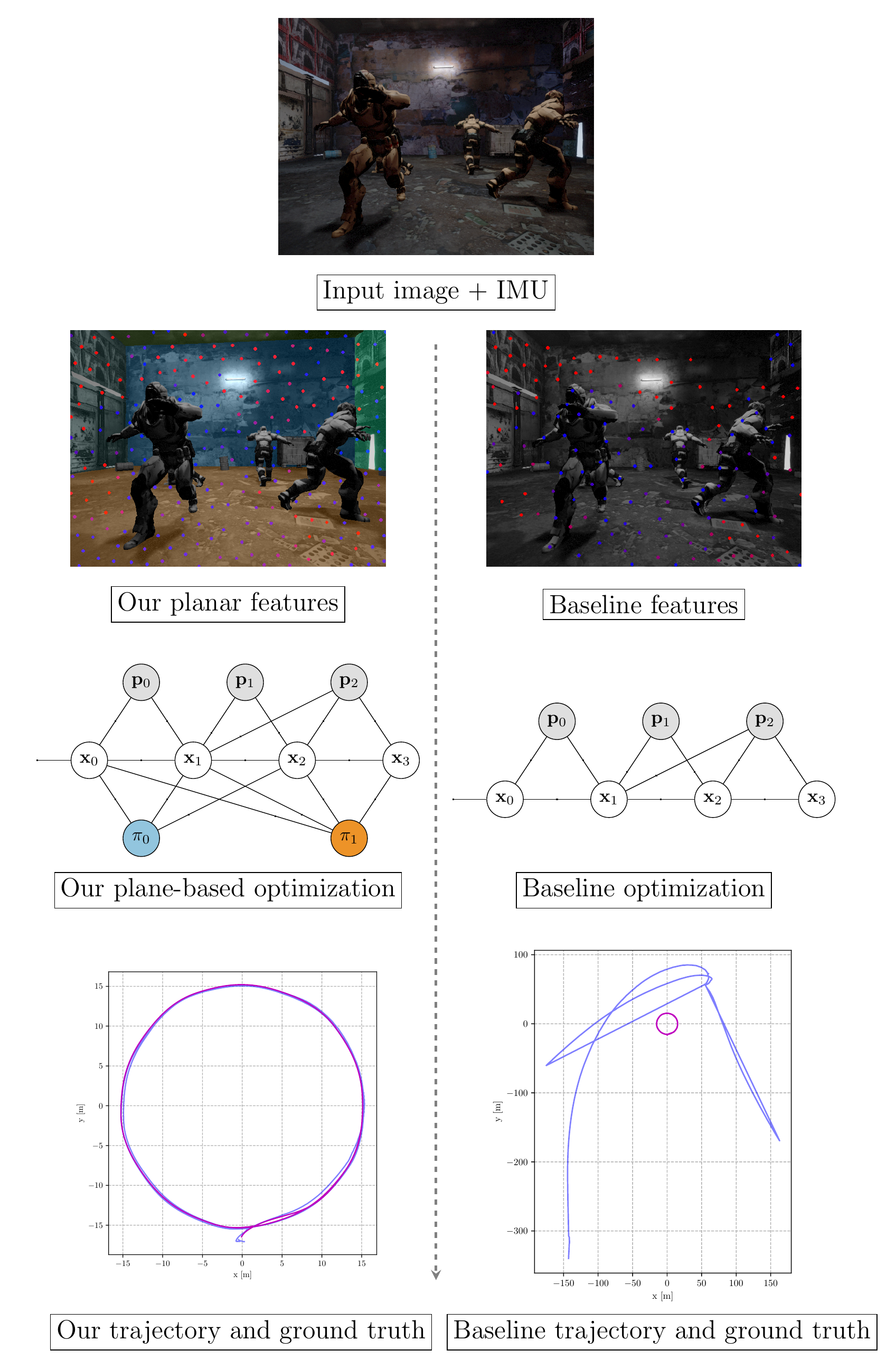}
    \caption{\textbf{Overview}: Motivated by the presence of large planar surfaces in man-made environments, we propose a monocular VIO system that estimates motion only from one or more planes in the scene based on their induced homographies, and ignoring all off-the-plane features. We show that this leads to improved robustness and accuracy in dynamic environments. {\color{blue}---} (blue) indicates the estimated trajectory and {\color{magenta} ---} (magenta) indicates the ground truth.}
    \vspace{-2em}
    \label{fig:teaser}
\end{figure}
Another significant limitation is their performance in visually dynamic environments that have multiple independently moving objects. The fundamental multiview geometry \cite{invitation3d} constraints hold only for static points and lead to errors when applied on dynamic points. This problem is especially significant during the initialization phase of monocular VINS, where pose estimates from visual SfM are usually directly aligned with the preintegrated IMU measurements to initialize the scale and IMU parameters. Incorrect visual pose estimates at this stage can lead to complete tracking failure, as we demonstrate in our experiments. The traditional approach of applying RANSAC to filter the dynamic features using a fundamental matrix model works well only for a small number of dynamic features, provided the features are also not following any degenerate motion profiles along the epipolar plane \cite{kundu-moving}. Motion segmentation approaches to directly predict the motion status of each image pixel often have elaborate multi-stage pipelines to distinguish the ego-motion from the object motion, resulting in high computation times \cite{motion-seg} that are not yet suitable for real-time SLAM systems.
 
Current best approaches use semantic labels to filter out features from potentially-dynamic semantic classes \cite{viode,ds-slam}.
Semantic segmentation, fuelled by deep learning, has seen tremendous progress and can produce accurate semantic labels at frame rate.
But the notion of a \emph{dynamic} class is handcrafted, and enumerating all possible dynamic classes leads to intractability. A tractable approach would be to instead directly identify static structures in a scene for feature tracking, bypassing semantics. We note that planar surfaces are the most abundant static regions in everyday man-made environments. Crucially, planes also offer a simple geometry that can be further exploited for an improved estimation. With this insight, we propose RP-VIO, a plane-based monocular visual-inertial odometry (VIO) system that is tailored for dynamic environments.

RP-VIO uses only features from one
or more planes in the scene, identified by a plane segmentation model, and uses the plane-induced homographies \cite{invitation3d} for motion estimation.
We augment a state-of-the-art monocular VIO system~\cite{vins-mono} with our proposed homography constraints, and significantly improve performance over the state-of-the-art on an in-house photorealistic synthetic dataset, as well as three diverse sequences from standard datasets including two real-world dynamic sequences.

 To summarize, our main contributions are as follows:
 \begin{itemize}
      \item RP-VIO, a monocular VIO system (built atop VINS-Mono~\cite{vins-mono}) that only uses planar features, and their induced homographies during both initialization and sliding-window estimation for improved robustness and accuracy in dynamic environments.
      \item A photorealistic visual-inertial dataset, which unlike existing datasets, contains dynamic characters present throughout the sequences (including initialization), and with sufficient IMU excitation.
      \item An extensive evaluation of our method against \cite{vins-mono} on our in-house dataset, an outdoor simulated sequence from the recently released VIODE dataset\cite{viode}, as well as two challenging real-world sequences from OpenLORIS-Scene\cite{openloris} and ADVIO\cite{advio} using a CNN-based plane segmentation model.
 \end{itemize}

\section{Related Work}
    \subsection{Visual-Inertial Odometry}

A concise overview of VINS research can be found in~\cite{vins-review}. We focus on closely related visual-inertial odometry algorithms (VIO) herein, which unlike visual-inertial SLAM systems~\cite{orb3,kimera}, only estimate the trajectory of the device and do not build a globally consistent map.

Filtering-based approaches to VIO still continue to be widely prevalent because of their efficiency.
An important work in this area is the multi-state constraint Kalman filter (MSCKF)~\cite{msckf} which adopts a structureless approach and marginalizes out the landmark positions, avoiding the quadratic EKF cost. A modern, performant implementation can be found in OpenVINS~\cite{openvins}.

Optimization-based approaches instead solve for the entire trajectory~\cite{vins-sam, dpi} or a sliding window of recent poses~\cite{okvis, vins-mono}, and are generally more accurate and robust.
An important enabler for optimization-based VINS is IMU preintegration~\cite{rot-preintegration} which allows multiple inertial measurements to be summarized, reducing state space size.

\subsection{Monocular VIO using Planes}

Plane-based visual-inertial systems that use stereo or depth sensors have been proposed in many works~\cite{dpi,struct-reg,point-plane-rgbd}. In monocular VIO systems however, planes are harder to segment accurately and their depths are also not directly available.

A monocular VIO system that uses only ground plane features, within an UKF was proposed in~\cite{plane-observ}. They also showed that the translation in the direction of the ground-plane normal becomes globally observable, reducing the total number of unobservable directions to three.
A direct frame-to-frame planar homography based VIO formulation was proposed in~\cite{rad-vio} for a downward-facing camera, but assumed a laser rangefinder for accurately estimating the scale.
A recent optimization-based monocular VIO system used an efficient plane and line parameterization~\cite{point-line}, while also leveraging a deep neural network for plane instance segmentation. However, all of these approaches have only been evaluated in static environments.

\subsection{VIO in Dynamic Environments}

A systematic survey of approaches for visual SLAM and visual odometry in dynamic environments can be found in~\cite{dynamic-slam-survey}.
Broadly, these approaches filter dynamic elements as outliers \cite{viode, ds-slam, motion-rgbd}, or jointly estimate the egomotion and the motion of the dynamic elements~\cite{vi-tracking, dynaslam2}. Our focus is on the former class of approaches as the latter approaches typically assume device egomotion to be readily estimated.

Relatively fewer approaches have specifically addressed VIO in dynamic environments. A method to detect conflicts between vision-only and inertial-only estimates has been presented in~\cite{motion-conflict}, but assuming the inertial measurements are always more reliable.
\cite{viode} exploited semantics to mask out dynamic objects for better egomotion estimation. However, this approach requires an enumeration of static and dynamic classes which is not always possible.

\subsubsection*{\textbf{Datasets}} \label{section:dynamic-datasets} The lack of publicly-available visual-inertial datasets that capture the dynamic nature of real-world environments has also made it difficult to evaluate the robustness of existing approaches. Progress has been made on this front with the recent release of the ADVIO\cite{advio} and OpenLORIS-Scene\cite{openloris} datasets. But the sequences in ADVIO are suitable only for a coarse, long-term evaluation of VIO algorithms since their ground truth is only sub-meter accurate. The sequences in OpenLORIS on the other hand were captured from a ground robot without sufficient excitation for the IMU which leads to unobservability\cite{vins-wheels}, making it difficult to isolate the effect of the dynamic characters. Recently, a challenging simulated dataset was proposed in \cite{viode}, along with an evaluation of two state-of-the-art VIO algorithms \cite{rovio, vins-mono} where they showed significant degradation of their accuracy. These sequences however do not contain enough dynamic characters present throughout the sequences, and during the initialization subsequence, which is the most fragile part of the system, there are no dynamic characters at all.

\textbf{Conclusion}: To the best of our knowledge, a monocular VIO system that optimizes over planar homographies and that is targeted at dynamic environments has not been proposed before. A fully dynamic visual-inertial dataset with accurate ground truth, synchronization, and sufficient observability also does not publicly exist.

\section{Method}
    While our proposed method is general enough to be integrated into any VIO or SLAM system, in this work we build upon VINS-Mono\cite{vins-mono}. VINS-Mono is a state-of-the-art, monocular VIO system that is based on a tightly-coupled sliding-window optimization of preintegrated IMU measurements and visual features. 
We consider it as a pure VIO system and ignore its relocalization and loop-closure modules. We build upon its front-end to detect and track only planar features in the scene, and introduce the induced planar homography constraints into its initialization and optimization modules.

\subsection{Definitions}

$\World$ denotes the world frame whose z-axis is in the downward direction along gravity. $\Imu$ denotes the body frame, which co-incides with the IMU frame, and $\Camera$ denotes the camera frame. $\Imu_i$ and $\Camera_i$ denote the body frame and camera frame at time $t_i$ respectively. $\R_{ji}$ and $\trans_{ji}$, together written as the homogeneous matrix $\T_{ji}$, denote the rotation and translation that transforms points from the frame at $t_i$ to the frame at $t_j$. The frame can be the camera frame or the body frame, depending on the context. $\R_i$ and $\trans_i$ denote the rotation and translation of the frame at $t_i$ with respect to the world frame. $\px^l$ denotes the normalized $2$D image coordinates of the $l$-th visual feature. The corresponding $3$D point $\pt_l$ is represented by its inverse depth $\lambda_l$ with respect to its first frame of observation. A plane $\boldsymbol{\pi}_p$ is represented by its normal and distance parameters $(\n, \depth)$ with respect to the $\Camera_0$ frame. The planar homography matrix (Fig. \ref{fig:homo}) which maps the $2$D image coordinates of a planar point from the $\Camera_0$ frame to the $\Camera_j$ frame is denoted as $\Homography_j$.

The state of our system at $t_i$, $\state_i$, is defined by the IMU position, orientation, velocity, biases, the inverse depth of the 3D features, and the plane parameters, i.e. $
\state_i \doteq [\R_i,\trans_i,\vel_i,\bias_i,\{\lambda_l\},\{\boldsymbol{\pi}_p\}]
$.

$\States$ denotes the state of all the frames within the sliding window $\mathcal{K}$, which we want to estimate, i.e. $\States \doteq \{\state_i\}_{i\in\mathcal{K}}$.

\subsection{Front-end}

Our system takes as input grayscale images, IMU measurements, and plane segmentation masks. These plane segmentation masks are obtained from a CNN-based model which we describe in Sec. \ref{section:plane-seg}. We apply the obtained plane instance segmentation masks on the original images to detect and track only the features that belong to the (static) planar regions in the scene, while also maintaining information about which plane each tracked feature belongs to. To avoid detecting any features along the edges of the mask which might belong to a dynamic object, we apply an erosion operation on the original masks. Further, we use RANSAC to fit a separate planar homography model to the features from each plane to discard any outliers. These outliers could be features arising from incorrect matches by the KLT optical flow algorithm, or from inaccurate segments that do not belong to the larger parent plane. The raw IMU measurements between image frames are converted into preintegrated measurements, and image frames with sufficient parallax and feature tracks are selected as keyframes.

\subsection{Initialization}

The main visual-inertial sliding-window optimization is non-convex and is minimized iteratively which requires an accurate initial estimate. To obtain a good initial estimate without making any assumptions about the starting configuration, a separate loosely-coupled initialization procedure is used where the visual measurements and inertial measurements are processed separately into their respective pose estimates and then aligned together to solve for the unknowns in multiple steps.

We begin by first solving for the camera poses, the $3$D points, and the plane parameters. From the window of initial image frames, two base frames having sufficient parallax are selected. Out of all their matching features we select only the ones that arise from the largest plane in the scene, i.e. the plane having the maximum number of features. Using these features, we fit a planar homography matrix $\Homography$ relating the two base frame poses and the largest plane using RANSAC. This homography matrix is normalized and then decomposed into the rotation, translation, and plane normal using the analytical method of Malis and Vargas \cite{hdecomp}, as implemented in OpenCV\cite{opencv}. The method however returns up to four different solution tuples which must be reduced to one. We first reduce this solution set to two by enforcing the positive depth constraint, i.e. all the plane features must lie in front of the camera. We implement this as the constraint, $\n_{i}^{\transpose}\px_{\mu} > 0$, where $\px_{\mu}$ is the mean $2$D feature point in normalized image coordinates. From the resulting two possible solutions, we finally select the one whose rotation (after transforming to $\Imu$ frame) is closest to the corresponding preintegrated IMU rotation $\mRot$,

\begin{equation}
\argmin_k\Vert \mRot^{\transpose}(\R_{\imu\camera}\R_{ij}^k\R_{\imu\camera}^{\transpose}) - \Identity \Vert^2
\end{equation}

Even though the gyroscope bias inside the preintegrated IMU rotation is not estimated yet, its magnitude is usually too small to cause a difference in the solution. The estimated pose from the decomposition is then used to triangulate the $3$D positions of the features between the two base frames and obtain an initial point cloud. The poses of the remaining frames within the window are estimated with respect to this point cloud using PnP. We note here that since the estimated pose between the two base frames is in the scale of the plane distance $\depth$, the triangulated point cloud and the deduced poses are also in the same scale. All the pose estimates are then fed into a visual bundle adjustment solver where, in addition to the standard $3$D-$2$D reprojection residual, we include the following $2$D-$2$D reprojection residual arising from the planar homography,

\begin{equation}
\residual_{\mathcal{H}} = \px_j^{l} - \left(\R_j + \frac{\trans_j\n^{\transpose}}{\depth}\right)\px^l
\end{equation}

\begin{figure}
\centering
\includegraphics[scale=0.8]{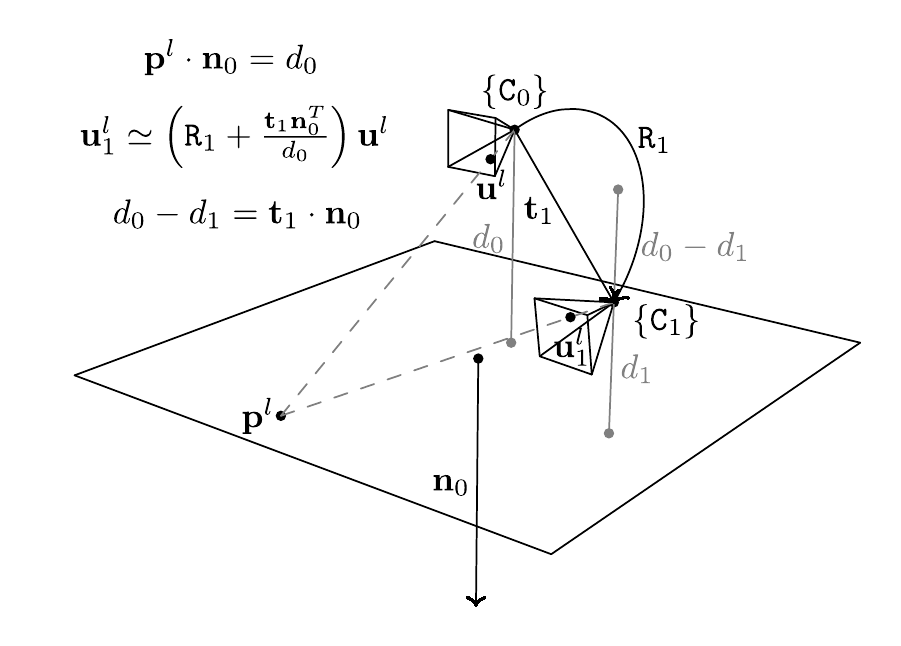}
\caption{The planar homography matrix $\Homography_1 = \left(\mathtt{R}_1 + \frac{\mathbf{t}_1\mathbf{n}^T_0}{d_0}\right)$ is a $3\times3$ matrix, arising from the planar constraint $\pt_l\cdot\n_0 = \depth_0$, which maps the observation $\px^l$ from the first frame to $\px_1^l$ in the second frame \cite{invitation3d}. The known ambiguity in its decomposition can be resolved using the IMU.}
\label{fig:homo}
\vspace{-1.5em}
\end{figure}

This residual measures the discrepancy between the expected observation of the point $\pt_l$ in frame $\Camera_j$, obtained by mapping its corresponding image location $\px_l$ from the first frame using the planar homography matrix, and the true observation $\px^l_j$. This is also illustrated in Fig. \ref{fig:homo}.
The output of this bundle adjustment is the up-to-scale ($\depth$) camera poses and $3$D feature points, and the plane normal. This unknown scale ($\depth$), along with the remaining unknowns needed to initialize the main optimization such as the gravity vector, velocities, and IMU biases are estimated using the same divide-and-conquer approach used in \cite{vins-mono}.

 Once these are estimated, the camera poses and $3$D feature points are re-scaled to metric units, and the world frame is re-aligned such that its Z-axis is in the direction of gravity. For the planes in the scene other than the largest plane, including planes that might be newly observed during operation, we similarly compute their respective planar homography matrices and decompose them. But for computation reasons, we avoid doing another round of bundle adjustment and the re-alignment of their poses with the IMU measurements to estimate the respective scale factors $\depth_p$. We instead directly estimate $\depth_p$ as the inverse ratio of each decomposed translation $\trans_p$ (which is in the scale of $\depth_p$ as $\frac{\trans}{\depth_p}$) to the corresponding metric translation $\trans$, which has already been estimated previously using the largest plane and inertial measurements. With this, all the visual and inertial quantities in our state have been solved for, and these estimates are fed into the sliding-window estimator as the initial seed for the optimization.

\subsection{Sliding-window Optimization}

A full batch optimization of the entire history of poses, map points, inertial and plane parameters quickly becomes computationally infeasible for real-time operation. Instead, a sliding-window of a fixed number of recent frames are optimized over their associated inertial and visual measurements. The optimization objective is described formally as follows.

\begin{figure}[H]
	\begin{center}
	    \includegraphics[scale=0.8]{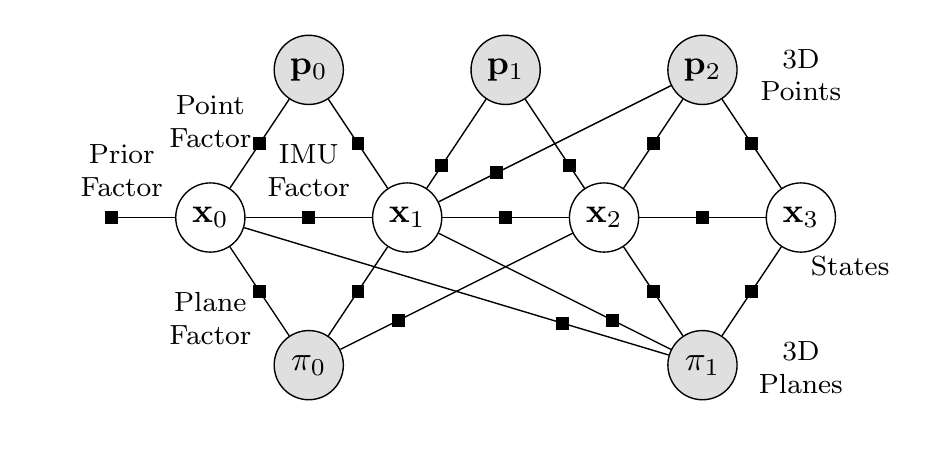}
		\caption{A factor graph representation of the (simplified) sliding window optimization showing the states and measurements linked together by the IMU, point, and plane factors.}
		\label{fig:factor-graph-structure}
	\end{center}
	\vspace{-1.5em}
\end{figure}

We denote with $\mathcal{I}_{ij}$ the set of all IMU measurements between two consecutive frame instances $i$ and $j$ within the window $\mathcal{K}$. The set of all planar features observed in frame $i$ is denoted as $\mathcal{C}_i$, and the set of all observed planes is denoted as $\mathcal{P}$. A factor graph representation of these states and measurements within a simplified window is shown in Fig. \ref{fig:factor-graph-structure}. The MAP estimate $\mathcal{X}^\star$ of all the states in the sliding window is obtained as the minimum of the sum of the squared residual errors,

\begin{align}
\States^\star &\doteq \arg\min_{\States}\;\left\Vert\residual_p\right\Vert^2 + \sum_{(i,j) \in \mathcal{K}}\left\Vert\residual_{\mathcal{I}_{ij}}\right\Vert^2 \nonumber
\\&
\quad + \sum_{i \in \mathcal{K}}\sum_{l \in \mathcal{C}_i}\rho\left(\left\Vert\residual_{\mathcal{C}_{il}}\right\Vert^2\right)
+ \sum_{p \in \mathcal{P}} \sum_{i \in \mathcal{K}} \sum_{l \in \mathcal{C}_i}  \rho\left(\left\Vert\residual_{\mathcal{H}}\right\Vert^2\right)
\end{align}

where $\residual_{p}$ is the prior residual resulting from marginalization of the previous states, $\residual_{\mathcal{I}_{ij}}$ is the preintegrated IMU residual, $\residual_{\mathcal{C}_{il}}$ is the standard $3$D-$2$D reprojection residual as defined in \cite{vins-mono}, and $\rho$ is a Cauchy loss that is used to down-weigh any outliers. $\residual_{\mathcal{H}}$ is our planar homography residual that is defined as,
\begin{equation}
\residual_{\mathcal{H}} = \px_j^{l} - \T_{\imu\camera}^{-1}\left(\R_{ji} + \frac{\trans_{ji}\n^{p^{\transpose}}_i}{\depth^p_i}\right)\T_{\imu\camera}\px_i^l
\end{equation}

This term is similar to the one used in the initialization, except the pose and plane parameters are in the body frame. The $p$-th plane normal $n^p$ and depth $\depth^p$ which are both originally defined in the first camera frame $\Camera_0$ are transformed to the current body frame $\Imu_i$ as follows,

\begin{align}
\n_{\imu_i} &= \R_{i}^{\transpose}\R_{\imu\camera}\n_{\camera_0} \nonumber \\
\depth_{\imu_0} &= \depth_{\camera_0} + \trans_{\imu\camera}\cdot\n_{\imu_0}\nonumber \\
\depth_{\imu_i} &= \depth_{\imu_0} - \trans_{\imu_{0}\imu_{i}}\cdot{\n_{\imu_0}}
\end{align}

\vspace{-2em}

This entire non-linear objective function is minimized iteratively using the Dogleg algorithm with Dense-Schur linear solver implemented in Ceres Solver\cite{ceres-solver}. At the end of the optimization, the window is moved forward by one frame to incorporate the latest frame. The state of the latest frame is initialized by propagating the inertial measurements from the previous frame. The dropped frame is marginalized as done in \cite{vins-mono}. 
The optimized plane parameters however are not dropped or marginalized and are instead reused as and when the plane is observed again.

\subsection{Plane Segmentation}
\label{section:plane-seg}

To segment the plane instances from each input RGB image, we use the Plane-Recover\cite{yang2018recovering} model. Their model is trained using a structure-induced loss to simultaneously predict plane segmentation masks and their $3$D parameters, with only semantic labels and no explicit $3$D annotations. The model runs on a single Nvidia GTX Titan X (Maxwell) GPU at $30$ FPS which also makes it suitable for real-time VIO.

Despite the effectiveness of their model, we noticed in our experiments that the predicted segments are often not continuous and single large planes were segmented as multiple separate planes. To overcome this we introduce an additional inter-plane loss function that constrains planes with small relative orientations between them into a single plane.

\begin{equation}
    \mathcal{L}_{\text{inter}} = \frac{1}{n}\sum_{i =1}^{n}\sum_{j=1}^{m} \Vert\n_{i}^{j}\cdot\n_{i}^{j} - l_i\Vert^2 
\end{equation}

where $\n$ is the plane normal, $m$ is the number of planes which we fix to $3$, $n$ is the number of images in a batch, and $l_i$ is the inter-plane label generated online that is assigned the value $1$ if $\angle(\n_i^j,\n_i^j) < \frac{\pi}{4}$ and $0$ otherwise.

\begin{figure*}[h]%
    \centering
    \subfloat{{\includegraphics[width=3cm]{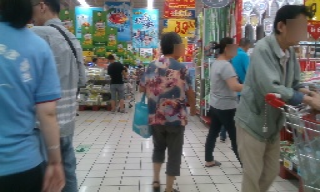}}}%
    \qquad
    \subfloat{{\includegraphics[width=3cm]{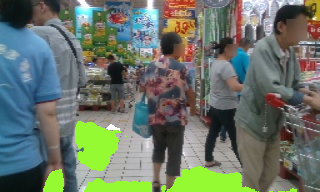} }}%
    \qquad
    \subfloat{{\includegraphics[width=3cm]{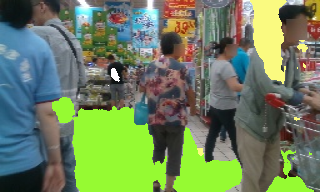} }}%
    \qquad
    \subfloat{{\includegraphics[width=3cm]{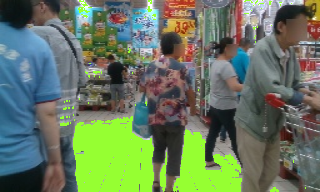} }}%
    \qquad \\ \vspace{10px}
    \subfloat[Inputs]{{\includegraphics[width=3cm]{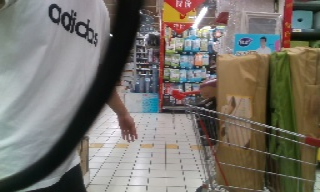} }}%
    \qquad
    \subfloat[PlaneRecover]{{\includegraphics[width=3cm]{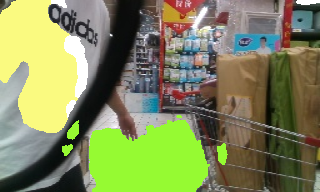} }}%
    \qquad
    \subfloat[Modified PlaneRecover]{{\includegraphics[width=3cm]{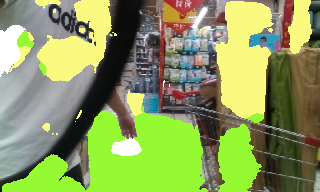} }}%
    \qquad
    \subfloat[CRF Refinement]{{\includegraphics[width=3cm]{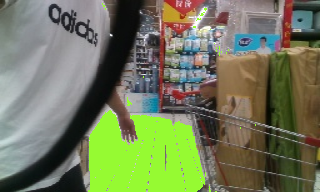} }}%
    \caption{Segmentation results for two challenging images in the OpenLORIS market-1 sequence, from the original model and the model modified with our inter-plane loss are visualized. The output from our model is further refined with a dense-CRF, before being used by our VIO system, and is also visualized here.}%
    \label{fig:seg-results}%
    \vspace{-1.5em}
\end{figure*}

With this added loss function, we retrain the network with their provided training data from SYNTHIA, and we train on two additional sequences ($00$, $01$) from the indoor ScanNet dataset. To further improve the segmentation and capture the fine boundary details, we employ a fully dense conditional random field (CRF) model \cite{CRF} that refines the network's segmentations. We use its default parameters as such without much tuning. Segmentation results from the model for an unseen real-world sequence that we use in our evaluations are visualized in Fig. \ref{fig:seg-results}.

To summarize, in this section we've described how we detect and track planar features from the scene, how we decompose their induced planar homography matrices into their respective motion and plane estimates using the IMU, and how we introduce the plane parameters as added constraints into the initializer and the sliding-window optimization. In the next section we demonstrate the effectiveness of this approach in dynamic environments.
\section{Experiments}
    We demonstrate using both simulated data and real-world data that using only planar features and their induced planar homography constraints leads to an improvement in estimation accuracy in dynamic environments. All the evaluations are run on a $6$-core Intel Core i$5$-$8400$ CPU with $8$ GB RAM and a $1$ TB HDD. To account for randomness from RANSAC and the multi-tasking OS, we report the median results from five runs for each evaluation. All code and data to reproduce our results are available on our project page.

\subsection{Simulation Experiments}

\subsubsection*{\textbf{RPVIO-Sim Dataset}}
For reasons explained in Sec. \ref{section:dynamic-datasets}, we generate our own dataset in simulation with accurate sensors and ground truth trajectories, and with sufficient IMU excitation throughout the sequences. We progressively add dynamic elements to these sequences and keep them visible in all parts of the sequences, even during initialization. This allows us to isolate their effect on the overall system accuracy.

We build a custom indoor warehouse environment with dynamic characters in Unreal Engine\cite{unrealengine}. We borrow several high-quality and feature-rich assets from the FlightGoggles\cite{flightgoggles} project for photorealism. This environment is integrated with AirSim\cite{airsim} to spawn a quadrotor and collect visual-inertial data. We collect monocular RGB images and their plane instance masks at $20$ Hz, IMU measurements and ground truth poses at $1000$ Hz. The IMU measurements are sub-sampled to $200$ Hz for our experiments. The camera and IMU intrinsics, and the camera-IMU spatial transform are obtained directly from AirSim. A time-offset of $0.03$ s between the camera and IMU measurements, introduced by the recording process, is calibrated using Kalibr\cite{kalibr}. 

The quadrotor is controlled to move along a circle of radius $15$ m, while moving along a sine wave in the vertical direction, resulting in a sinusoidal pattern. The sine excitation along the height is to ensure a non-constant acceleration and keep the scale observable \cite{vins-wheels}. We further command it to accelerate vertically at the beginning of its motion, before following the trajectory, to help the initialization. The total trajectory is of $200$ m length and $80$ s duration, with a maximum speed of $3$ m/s. Within the circle formed by the quadrotor, we introduce dynamic characters that are performing a repetitive dance motion. We progressively add more dynamic characters to each sequence, keeping everything else fixed, starting from no characters (static) and going up to $8$ characters (C$8$), recording six sequences in total. The yaw-direction of the quadrotor is also fixed to keep the camera pointing towards the center of the circle, such that the characters are in the FoV of the camera for the entire sequence. The quadrotor and the characters are controlled programmatically to ensure their motions are repeatable and are in sync across all the sequences.

\definecolor{dark-green}{RGB}{0,136,55}
\begin{figure}[t]
\centering
    \begin{subfigure}[t]{0.45\textwidth}
        \centering
        \includegraphics[width=0.9\textwidth]{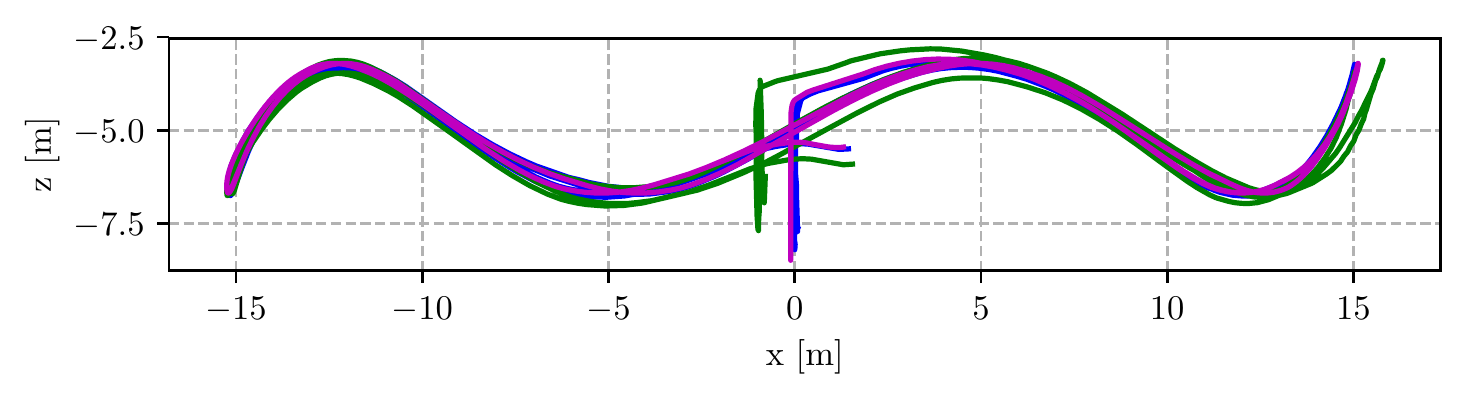}
        \caption{Side-view comparison between VINS-Mono {\color{dark-green}\textbf{---}} (green), RPVIO-Single {\color{blue}\textbf{---}} (blue), and ground truth {\color{magenta}\textbf{---}} (magenta) on the C$2$ sequence. VINS-Mono accumulates error during its initialization (vertical motion), while RPVIO-Single tracks accurately throughout the sequence.}
        \label{fig:c2-side}
        \vspace{0.5em}
    \end{subfigure}
    \begin{subfigure}[t]{0.45\textwidth}
        \centering
        \includegraphics[width=0.6\textwidth]{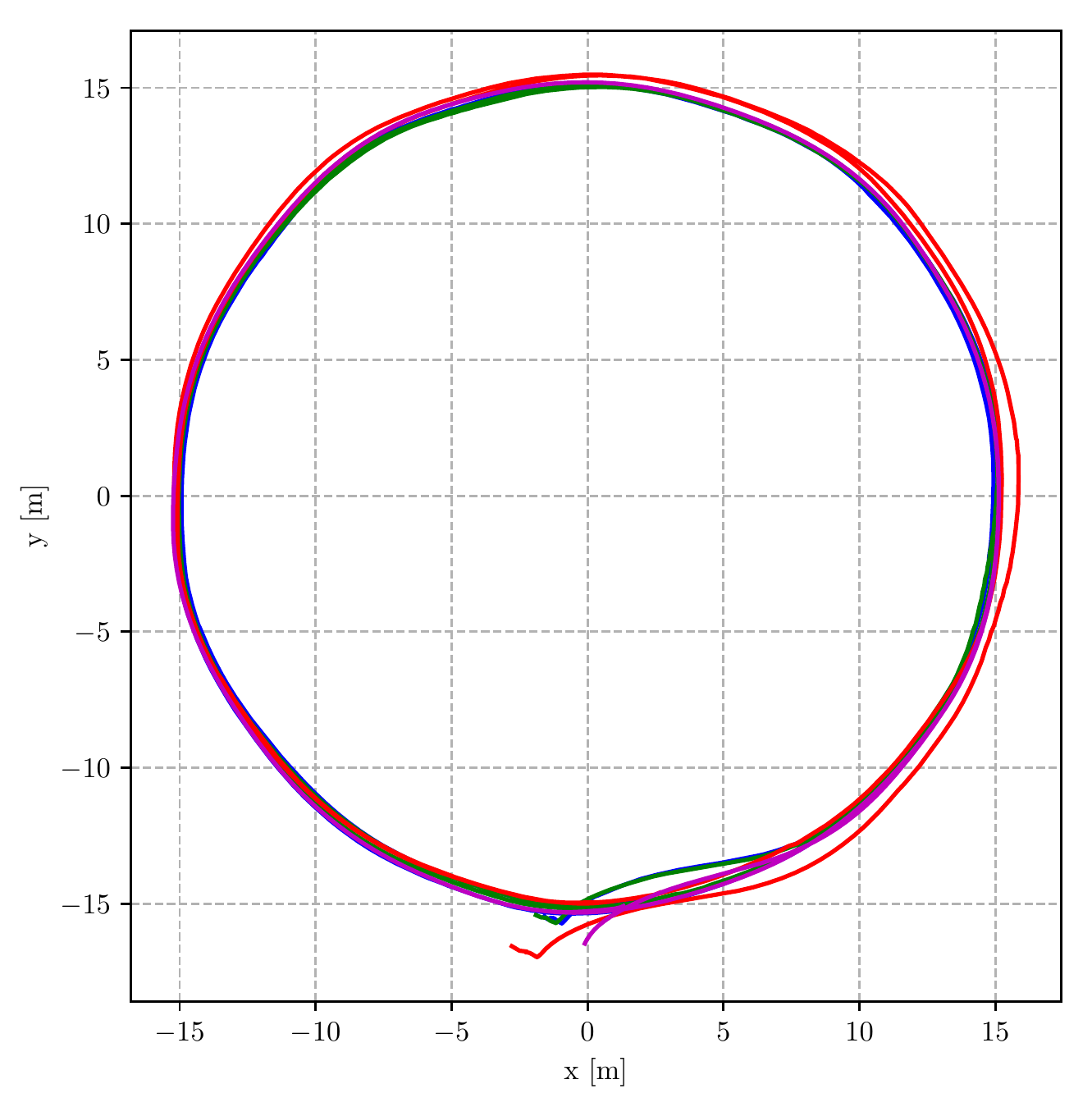}
        \caption{Top-view comparison between Mask-VINS {\color{red}\textbf{---}} (red), RPVIO-Single {\color{blue}\textbf{---}} (blue), RPVIO-Multi {\color{dark-green}\textbf{---}} (green), and ground truth {\color{magenta}\textbf{---}} (magenta) on the C$6$ sequence. Original VINS-Mono fails to track completely and is not included. Both RPVIO-Single and RPVIO-Multi are closer to the ground truth than Mask-VINS.}
        \label{fig:c6-top}
    \end{subfigure}
    \caption{Trajectory comparisons from our simulated experiments.}
    \vspace{-1.5em}
    \end{figure}

\subsubsection*{\textbf{Evaluation}} We evaluate VINS-Mono\cite{vins-mono}, and our proposed method on these generated sequences. We use two versions of our method, RPVIO-Single and RPVIO-Multi. RPVIO-Single includes in the optimization only features from the largest plane visible at any time, while RPVIO-Multi includes features from all the visible planes. We also create another version of VINS-Mono, called Mask-VINS, that is modified to take as an additional input the same plane instance masks as ours. It uses these masks to detect and track all the features that belong to all the static planar regions in the environment while avoiding features from all the dynamic characters, similar to \cite{viode}. It uses the same feature parameters as ours, and the masks are also eroded to avoid tracking features along the mask edges which might belong to dynamic characters. The back-end remains the same as VINS-Mono. We use this additional version to investigate the effect of the added planar homography residual term $\residual_{\mathcal{H}}$ in the optimization. We compute the RMSE of the estimated trajectories of each method for every sequence, after \SEthree alignment\cite{eval-tut} with the ground truth trajectories, and report them in Tab. \ref{table:sim-results}.

\begin{table}[h]
\caption{Results of the evaluation on our simulated dataset. We report the median RMSE from five runs on each sequence. X denotes complete tracking failure. Results which show a significant improvement are underlined.}
\centering
\ra{1.3}
\begin{tabular}{lcccc}
    \toprule
        {} & \multicolumn{4}{c}{Absolute Trajectory RMSE (m)}\\
        \cmidrule{2-5}
        Seq. & VINS-Mono & Mask-VINS & RPVIO-Multi & RPVIO-Single\\
    \midrule
        Static & $0.21$ & - & $\mathbf{0.19}$ & $\mathbf{0.19}$\\
        C$1$ & $0.24$ & $\mathbf{0.23}$ & $0.28$ & $\mathbf{0.23}$\\
        C$2$ & $0.85$ & $0.21$ & $0.24$ & $\mathbf{0.18}$\\
        C$4$ & X & $0.68$ & $0.76$ & $\underline{\mathbf{0.56}}$\\
        C$6$ & X & $0.91$ & $0.62$ & $\underline{\mathbf{0.54}}$\\
        C$8$ & X & X & $\underline{\mathbf{0.77}}$ & $0.85$\\
    \bottomrule
\end{tabular}
\label{table:sim-results}
\end{table}

\subsubsection*{\textbf{Discussion}}
The performance of VINS-Mono, Mask-VINS, and RP-VIO on the static and one character sequences are very similar. Since the number of static points are much greater than the number of dynamic points, the effect of RANSAC is the same as applying the mask. In the two character sequence we note that VINS-Mono has a much lower accuracy than Mask-VINS and RP-VIO, while the accuracy of Mask-VINS and RP-VIO are again similar. VINS-Mono accumulates most of the error during the initialization as shown in \ref{fig:c2-side}, when one of the characters is close to the camera. In the four, six, and eight character sequences however, VINS-Mono's initialization error is too high and it loses track completely. Mask-VINS and RP-VIO are still able to track successfully in C$4$ and C$6$, but RPVIO-Single is the most accurate (also shown in Fig. \ref{fig:c6-top}) which alludes to the role of the added homography constraints in the improved robustness. In the C$8$ sequence ours is still able to track successfully like the other sequences but Mask-VINS loses track completely. This could be because the scene is very cluttered and the few features that are left come only from a single plane during initialization which is a degenerate case for VINS-Mono's fundamental matrix based SfM initializer. RPVIO-Multi shows a better accuracy than RPVIO-Single in this sequence which could be because unlike in the previous sequences RPVIO-Single has fewer stable features to track than RPVIO-Multi.

\subsection{Experiments on Standard Datasets}

\captionsetup[subfloat]{labelformat=empty}

\subsubsection*{\textbf{Sequences}}
We evaluate the robustness of our system on three more sequences from three diverse datasets. The first sequence is from the newly released VIODE\cite{viode} dataset, that was also generated using AirSim. This sequence was captured in an outdoor city environment with many moving vehicles, from a drone that is performing very aggressive maneuvers including sharp rotations. We use their provided segmentation masks to track features only along the road. The second sequence is from the OpenLORIS-Scene\cite{openloris} dataset. This was captured in a real-world supermarket from a floor-cleaning robot that contains many dynamic characters in the form of moving people, trolleys, and bags. The third sequence is from the ADVIO\cite{advio} dataset that was captured from a hand-held smartphone in a real-world metro station. This is the most visually challenging sequence out of the three, with a narrow FoV and fast motions, and dynamic characters in the form of a large moving train and people. The total lengths of the three sequences are $166$ m, $145$ m, and $136$ m respectively.

\subsubsection*{\textbf{Evaluation}}
We use RPVIO-Single for all the three sequences since they contain predominantly only a single large plane that can be tracked reliably at a time. We use the same feature parameters that were used for the simulation experiments without any tuning. We compute the RMSE ATE of its estimated trajectories with respect to the ground truth after $\SEthree$ alignment and compare against VINS-Mono. The median errors from five runs for each sequence are reported in Tab. \ref{table:std-results}. Since all the masked features predominantly come from a single plane in all the three sequences, we do not compare against Mask-VINS that was used in the earlier evaluation since features from a single plane form a degenerate configuration for the original VINS-Mono initializer. Further, the unavailability of an off-the-shelf semantic classifier that can accurately segment all the dynamic objects present in both the real-world sequences also makes a fair comparison with the Mask-VINS approach not possible. The images in the ADVIO sequence are of a very high resolution $1280\times720$ and come at a high rate of $60$ Hz which causes a lot of frame drops in the VINS front-end. For this reason the evaluation on this sequence alone is run on a $2$ GHz $12$-core Intel Xeon CPU with $32$GB RAM and an SSD.

\begin{table}[h]
\caption{Results of the evaluations on three diverse sequences. We report the median RMSE from five runs on each sequence.}
\centering
\ra{1.3}
\begin{tabular}{lcc}
    \toprule
        {} & \multicolumn{2}{c}{Absolute Trajectory RMSE (m)}\\
        \cmidrule{2-3}
        Sequence & VINS-Mono & RPVIO-Single\\
    \midrule
        VIODE-city-night-high & $0.73$ & $\mathbf{0.32}$\\
        OpenLORIS-market-$1$ & $2.45$ & $\mathbf{1.35}$\\
        ADVIO-$12$ & $4.34$ & $\mathbf{2.75}$\\
    \bottomrule
\end{tabular}
\label{table:std-results}
\end{table}

\subsubsection*{\textbf{Discussion}}

Our method shows a significant improvement over VINS-Mono on all three sequences. In the OpenLORIS and VIODE sequences, our method used a lesser number of features than VINS-Mono despite which it has shown greater accuracy. This makes us believe that it might be sufficient to track few stable features than tracking all possible features, of which many can be noisy.
On both the real-world sequences, despite using a generic plane detection network that has not been re-trained, the network and the CRF are able to provide reliable plane segmentations that are still good enough for our method to track accurately. If scene-specific training data is available, we expect more accurate segmentations and a better overall trajectory estimate. For scenes which can contain dynamic planar surfaces such as from vehicles, specific ground or wall surface classifiers must be trained and used instead. Training such specific surface classifiers is still a more feasible approach than trying to train semantic classifiers to segment all possible moving objects. In the absence of clear planar structures however, our method should be considered complimentary to general-purpose point-based systems as and when planes become visible, and not as a complete replacement. %

\begin{figure}
    \centering
    \includegraphics[scale=0.5]{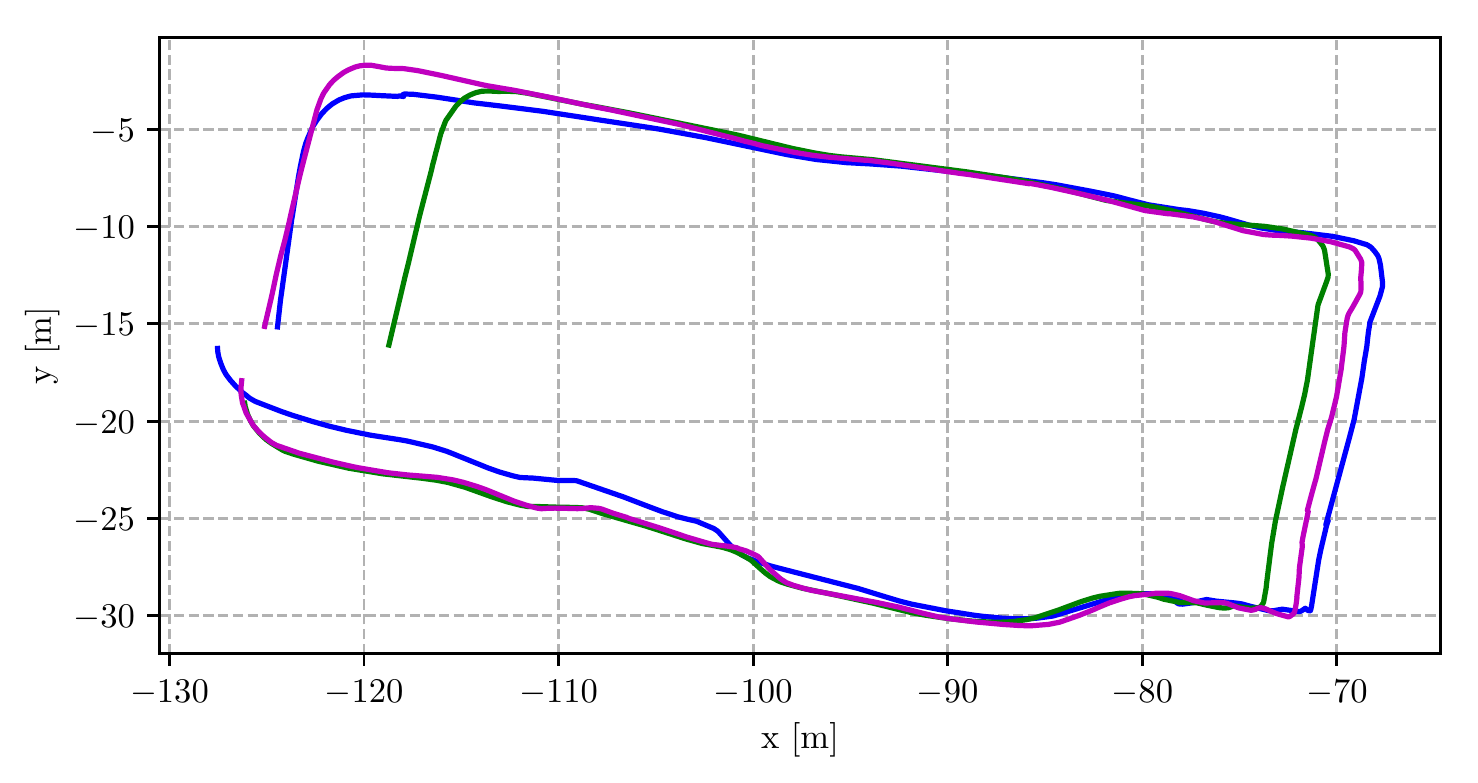}
    \caption{Top-view comparison between RPVIO-Single {\color{blue}\textbf{---}} (blue), VINS-Mono {\color{dark-green}\textbf{---}} (green), and ground truth {\color{magenta}\textbf{---}} (magenta), on the OpenLORIS market-1 sequence.}
    \label{fig:ol-result}
    \vspace{-1.5em}
\end{figure}

\section{Conclusion}
    We have proposed a monocular VIO system that uses only one or more planes in the environment and their structural regularity for an accurate motion estimation in dynamic environments. We have validated its improved performance in diverse simulated and real-world dynamic environments, while showing the same baseline performance in static environments. For real-world environments, using only a generic plane segmentation model, we showed an improvement of up to $45\%$ over a state-of-the-art monocular VIO system. We have also shown in our comparison with Mask-VINS in simulation that our approach achieves better accuracy than a simple dynamic-features masking approach which also alludes to the role of the added structural constraints in the improved robustness. The future scope of this work is to extend RP-VIO into a full SLAM system to obtain clean and consistent plane-based maps, without any off-the-plane noisy features. Such an approach can also make use of additional Manhattan constraints and even corresponding line features for improved accuracy. Further, it can be investigated if the predicted $3$D plane parameters from the plane segmentation model can be used for a faster initialization.

\hypersetup{
    urlcolor=black
    }

\bibliography{references}

\begin{thebibliography}{10}
\providecommand{\url}[1]{#1}
\csname url@samestyle\endcsname
\providecommand{\newblock}{\relax}
\providecommand{\bibinfo}[2]{#2}
\providecommand{\BIBentrySTDinterwordspacing}{\spaceskip=0pt\relax}
\providecommand{\BIBentryALTinterwordstretchfactor}{4}
\providecommand{\BIBentryALTinterwordspacing}{\spaceskip=\fontdimen2\font plus
\BIBentryALTinterwordstretchfactor\fontdimen3\font minus
  \fontdimen4\font\relax}
\providecommand{\BIBforeignlanguage}[2]{{%
\expandafter\ifx\csname l@#1\endcsname\relax
\typeout{** WARNING: IEEEtran.bst: No hyphenation pattern has been}%
\typeout{** loaded for the language `#1'. Using the pattern for}%
\typeout{** the default language instead.}%
\else
\language=\csname l@#1\endcsname
\fi
#2}}
\providecommand{\BIBdecl}{\relax}
\BIBdecl

\bibitem{eth-nav}
H.~Oleynikova, C.~Lanegger \emph{et~al.}, ``An open-source system for
  vision-based micro-aerial vehicle mapping, planning, and flight in cluttered
  environments,'' \emph{Journal of Field Robotics}, vol.~37, no.~4, pp.
  642--666, 2020.

\bibitem{oculus}
Facebook, ``Oculus vr,'' \url{https://oculus.com}.

\bibitem{hololens}
Microsoft, ``Hololens,'' \url{https://microsoft.com/en-us/hololens}.

\bibitem{vins-wheels}
K.~J. Wu, C.~X. Guo, G.~Georgiou, and S.~I. Roumeliotis, ``{VINS} on wheels,''
  in \emph{2017 {IEEE} International Conference on Robotics and Automation
  ({ICRA})}, May 2017, pp. 5155--5162.

\bibitem{degenerate-ex}
Y.~Yang, P.~Geneva, K.~Eckenhoff, and G.~Huang, ``Degenerate motion analysis
  for aided ins with online spatial and temporal sensor calibration,''
  \emph{IEEE Robotics and Automation Letters}, vol.~4, no.~2, pp. 2070--2077,
  2019.

\bibitem{degenerate-in}
Y.~Yang, P.~Geneva, X.~Zuo, and G.~Huang, ``Online {IMU} intrinsic calibration:
  Is it necessary?'' in \emph{Robotics: Science and Systems {XVI}}.\hskip 1em
  plus 0.5em minus 0.4em\relax Robotics: Science and Systems Foundation, Jul.
  2020.

\bibitem{invitation3d}
Y.~Ma, S.~Soatto, J.~Kosecka, and S.~S. Sastry, \emph{An invitation to 3-d
  vision: from images to geometric models}.\hskip 1em plus 0.5em minus
  0.4em\relax Springer Science \& Business Media, 2012, vol.~26.

\bibitem{kundu-moving}
A.~Kundu, K.~M. Krishna, and J.~Sivaswamy, ``Moving object detection by
  multi-view geometric techniques from a single camera mounted robot,'' in
  \emph{2009 IEEE/RSJ International Conference on Intelligent Robots and
  Systems}.\hskip 1em plus 0.5em minus 0.4em\relax IEEE, 2009, pp. 4306--4312.

\bibitem{motion-seg}
J.~Vertens, A.~Valada, and W.~Burgard, ``{SMSnet}: Semantic motion segmentation
  using deep convolutional neural networks,'' in \emph{2017 {IEEE/RSJ}
  International Conference on Intelligent Robots and Systems ({IROS})}, Sep.
  2017, pp. 582--589.

\bibitem{viode}
K.~{Minoda}, F.~{Schilling} \emph{et~al.}, ``Viode: A simulated dataset to
  address the challenges of visual-inertial odometry in dynamic environments,''
  \emph{IEEE Robotics and Automation Letters}, pp. 1--1, 2021.

\bibitem{ds-slam}
C.~Yu, Z.~Liu \emph{et~al.}, ``{DS-SLAM}: A semantic visual {SLAM} towards
  dynamic environments,'' in \emph{2018 {IEEE/RSJ} International Conference on
  Intelligent Robots and Systems ({IROS})}, Oct. 2018, pp. 1168--1174.

\bibitem{vins-mono}
T.~Qin, P.~Li, and S.~Shen, ``{VINS-Mono}: A robust and versatile monocular
  {Visual-Inertial} state estimator,'' \emph{IEEE Trans. Rob.}, vol.~34, no.~4,
  pp. 1004--1020, Aug. 2018.

\bibitem{openloris}
X.~Shi, D.~Li \emph{et~al.}, ``Are we ready for service robots? the
  {OpenLORIS-Scene} datasets for lifelong {SLAM},'' in \emph{2020 International
  Conference on Robotics and Automation (ICRA)}, 2020, pp. 3139--3145.

\bibitem{advio}
S.~Cort{\'e}s, A.~Solin, E.~Rahtu, and J.~Kannala, ``{ADVIO}: An authentic
  dataset for visual-inertial odometry,'' in \emph{Proceedings of the European
  Conference on Computer Vision ({ECCV})}, 2018, pp. 419--434.

\bibitem{vins-review}
G.~Huang, ``Visual-inertial navigation: A concise review,'' in \emph{2019
  international conference on robotics and automation (ICRA)}.\hskip 1em plus
  0.5em minus 0.4em\relax IEEE, 2019, pp. 9572--9582.

\bibitem{orb3}
C.~Campos, R.~Elvira \emph{et~al.}, ``{ORB-SLAM3}: An accurate open-source
  library for visual, visual-inertial and multi-map {SLAM},'' \emph{arXiv
  preprint arXiv:2007.11898}, 2020.

\bibitem{kimera}
\BIBentryALTinterwordspacing
A.~Rosinol, M.~Abate, Y.~Chang, and L.~Carlone, ``Kimera: an open-source
  library for real-time metric-semantic localization and mapping,'' in
  \emph{IEEE Intl. Conf. on Robotics and Automation (ICRA)}, 2020. [Online].
  Available: \url{https://github.com/MIT-SPARK/Kimera}
\BIBentrySTDinterwordspacing

\bibitem{msckf}
A.~I. Mourikis and S.~I. Roumeliotis, ``A multi-state constraint kalman filter
  for vision-aided inertial navigation,'' in \emph{Proceedings 2007 IEEE
  International Conference on Robotics and Automation}.\hskip 1em plus 0.5em
  minus 0.4em\relax IEEE, 2007, pp. 3565--3572.

\bibitem{openvins}
P.~Geneva, K.~Eckenhoff \emph{et~al.}, ``Openvins: A research platform for
  visual-inertial estimation,'' in \emph{Proc. of the IEEE International
  Conference on Robotics and Automation}, Paris, France, 2020.

\bibitem{vins-sam}
V.~Indelman, S.~Williams, M.~Kaess, and F.~Dellaert, ``Factor graph based
  incremental smoothing in inertial navigation systems,'' in \emph{2012 15th
  International Conference on Information Fusion}.\hskip 1em plus 0.5em minus
  0.4em\relax IEEE, 2012, pp. 2154--2161.

\bibitem{dpi}
M.~Hsiao, E.~Westman, and M.~Kaess, ``Dense planar-inertial slam with
  structural constraints,'' in \emph{2018 IEEE International Conference on
  Robotics and Automation (ICRA)}.\hskip 1em plus 0.5em minus 0.4em\relax IEEE,
  2018, pp. 6521--6528.

\bibitem{okvis}
S.~Leutenegger, S.~Lynen \emph{et~al.}, ``Keyframe-based visual--inertial
  odometry using nonlinear optimization,'' \emph{The International Journal of
  Robotics Research}, vol.~34, no.~3, pp. 314--334, 2015.

\bibitem{rot-preintegration}
C.~Forster, L.~Carlone, F.~Dellaert, and D.~Scaramuzza, ``{On-Manifold}
  preintegration for {Real-Time} {Visual-Inertial} odometry,'' \emph{IEEE
  Trans. Rob.}, vol.~33, no.~1, pp. 1--21, Feb. 2017.

\bibitem{struct-reg}
A.~Rosinol, T.~Sattler, M.~Pollefeys, and L.~Carlone, ``Incremental
  visual-inertial 3d mesh generation with structural regularities,'' in
  \emph{2019 International Conference on Robotics and Automation (ICRA)}.\hskip
  1em plus 0.5em minus 0.4em\relax IEEE, 2019, pp. 8220--8226.

\bibitem{point-plane-rgbd}
Y.~{Yang}, P.~{Geneva} \emph{et~al.}, ``Tightly-coupled aided inertial
  navigation with point and plane features,'' in \emph{2019 International
  Conference on Robotics and Automation (ICRA)}, 2019, pp. 6094--6100.

\bibitem{plane-observ}
G.~Panahandeh, S.~Hutchinson, P.~H{\"a}ndel, and M.~Jansson, ``{Planar-Based}
  visual inertial navigation: Observability analysis and motion estimation,''
  \emph{J. Intell. Rob. Syst.}, vol.~82, no.~2, pp. 277--299, May 2016.

\bibitem{rad-vio}
B.~{Fu}, K.~S. {Shankar}, and N.~{Michael}, ``Rad-vio: Rangefinder-aided
  downward visual-inertial odometry,'' in \emph{2019 International Conference
  on Robotics and Automation (ICRA)}, 2019, pp. 1841--1847.

\bibitem{point-line}
X.~Li, Y.~Li \emph{et~al.}, ``Co-planar parametrization for stereo-slam and
  visual-inertial odometry,'' \emph{IEEE Robotics and Automation Letters},
  vol.~5, no.~4, pp. 6972--6979, 2020.

\bibitem{dynamic-slam-survey}
M.~R.~U. Saputra, A.~Markham, and N.~Trigoni, ``\BIBforeignlanguage{en}{Visual
  {SLAM} and structure from motion in dynamic environments: A survey},''
  \emph{\BIBforeignlanguage{en}{ACM Comput. Surv.}}, vol.~51, no.~2, pp. 1--36,
  Feb. 2018.

\bibitem{motion-rgbd}
Y.~Sun, M.~Liu, and M.~Q.~H. Meng, ``Improving {RGB-D} {SLAM} in dynamic
  environments: A motion removal approach,'' \emph{Rob. Auton. Syst.}, 2017.

\bibitem{vi-tracking}
K.~Eckenhoff, Y.~Yang, P.~Geneva, and G.~Huang, ``{Tightly-Coupled}
  {Visual-Inertial} localization and {3-D} {Rigid-Body} target tracking,''
  \emph{IEEE Robotics and Automation Letters}, vol.~4, no.~2, pp. 1541--1548,
  Apr. 2019.

\bibitem{dynaslam2}
B.~Bescos, C.~Campos, J.~D. Tardós, and J.~Neira, ``Dynaslam ii:
  Tightly-coupled multi-object tracking and slam,'' \emph{IEEE Robotics and
  Automation Letters}, vol.~6, no.~3, pp. 5191--5198, 2021.

\bibitem{motion-conflict}
B.~P.~W. Babu, D.~Cyganski, J.~Duckworth, and S.~Kim, ``Detection and
  resolution of motion conflict in visual inertial odometry,'' in \emph{2018
  {IEEE} International Conference on Robotics and Automation ({ICRA})}, May
  2018, pp. 996--1002.

\bibitem{rovio}
M.~Bloesch, M.~Burri \emph{et~al.}, ``Iterated extended kalman filter based
  visual-inertial odometry using direct photometric feedback,'' \emph{The
  International Journal of Robotics Research}, vol.~36, no.~10, pp. 1053--1072,
  2017.

\bibitem{hdecomp}
E.~Malis and M.~Vargas, ``Deeper understanding of the homography decomposition
  for vision-based control,'' INRIA, Tech. Rep. inria-00174036, 2007.

\bibitem{opencv}
G.~Bradski, ``{The OpenCV Library},'' \emph{Dr. Dobb's Journal of Software
  Tools}, 2000.

\bibitem{ceres-solver}
S.~Agarwal, K.~Mierle, and Others, ``Ceres solver,''
  \url{https://ceres-solver.org}.

\bibitem{yang2018recovering}
F.~Yang and Z.~Zhou, ``Recovering 3d planes from a single image via
  convolutional neural networks,'' in \emph{Proceedings of the European
  Conference on Computer Vision (ECCV)}, 2018, pp. 85--100.

\bibitem{CRF}
P.~Kr{\"a}henb{\"u}hl and V.~Koltun, ``Efficient inference in fully connected
  crfs with gaussian edge potentials,'' \emph{arXiv preprint arXiv:1210.5644},
  2012.

\bibitem{unrealengine}
E.~Games, ``Unreal engine,'' \url{http://unrealengine.com}.

\bibitem{flightgoggles}
W.~Guerra, E.~Tal \emph{et~al.}, ``{FlightGoggles}: Photorealistic sensor
  simulation for perception-driven robotics using photogrammetry and virtual
  reality,'' in \emph{2019 {IEEE/RSJ} International Conference on Intelligent
  Robots and Systems ({IROS})}.\hskip 1em plus 0.5em minus 0.4em\relax IEEE,
  Nov. 2019, pp. 6941--6948.

\bibitem{airsim}
S.~Shah, D.~Dey, C.~Lovett, and A.~Kapoor, ``{AirSim}: {High-Fidelity} visual
  and physical simulation for autonomous vehicles,'' in \emph{Field and Service
  Robotics}.\hskip 1em plus 0.5em minus 0.4em\relax Springer International
  Publishing, 2018, pp. 621--635.

\bibitem{kalibr}
P.~Furgale, J.~Rehder, and R.~Siegwart, ``Unified temporal and spatial
  calibration for multi-sensor systems,'' in \emph{2013 {IEEE/RSJ}
  International Conference on Intelligent Robots and Systems}, Nov. 2013, pp.
  1280--1286.

\bibitem{eval-tut}
Z.~Zhang and D.~Scaramuzza, ``A tutorial on quantitative trajectory evaluation
  for {Visual(-Inertial}) odometry,'' in \emph{2018 {IEEE/RSJ} International
  Conference on Intelligent Robots and Systems ({IROS})}, Oct. 2018, pp.
  7244--7251.

\end{thebibliography}
\footnotesize{
\bibliographystyle{IEEEtran}
}

\addtolength{\textheight}{-12cm}   %

\end{document}